\documentclass[nonatbib,preprint,12pt,authoryear]{elsarticle}

\usepackage{amssymb}
\usepackage{apacite}

\usepackage{todonotes} % for todos

\usepackage{tcolorbox}
\usepackage{framed}
\usepackage{xcolor}
\definecolor{mygray}{gray}{.9}
\definecolor{shadecolor}{rgb}{0.92,0.92,0.92}

\usepackage{CJKutf8}

\usepackage{colortbl}

\journal{Arxiv}

\begin{document}

\begin{frontmatter}



\title{Data Augmentation for Fake Reviews Detection in Multiple Languages and Multiple Domains}


% \author{}
\author{Ming Liu and Massimo Poesio}
\ead{\{acw661,m.poesio\}@qmul.ac.uk}
\affiliation{organization={Queen Mary University of London},%Department and Organization
            addressline={Mile End Road}, 
            city={London},
            postcode={E1 4NS}
            % country={}
            }

% \begin{abstract}
\begin{abstract}
With the growth of the Internet, buying habits have changed, and customers have become more dependent on the online opinions of other customers to guide their purchases. 
Identifying fake reviews thus became an important area for Natural Language Processing (NLP) research. 
However, developing high-performance NLP models depends on the availability of large amounts of training data, which are often not available for low-resource languages or domains. 
In this research, we used large language models to generate datasets to train fake review detectors. 
Our approach was used to generate fake reviews in different domains (book reviews, restaurant reviews, and hotel reviews) and different languages (English and Chinese). 
Our results demonstrate that our data augmentation techniques result in improved performance at fake review detection for all domains and languages. The accuracy of our fake review detection model can be improved by
0.3 percentage points on DeRev TEST, 
10.9 percentage points on Amazon TEST, 8.3 percentage points on Yelp TEST and 7.2 percentage points on DianPing TEST
using the augmented datasets.
\end{abstract}
% \end{abstract}

%Graphical abstract
\begin{graphicalabstract}
\includegraphics[width=\textwidth]{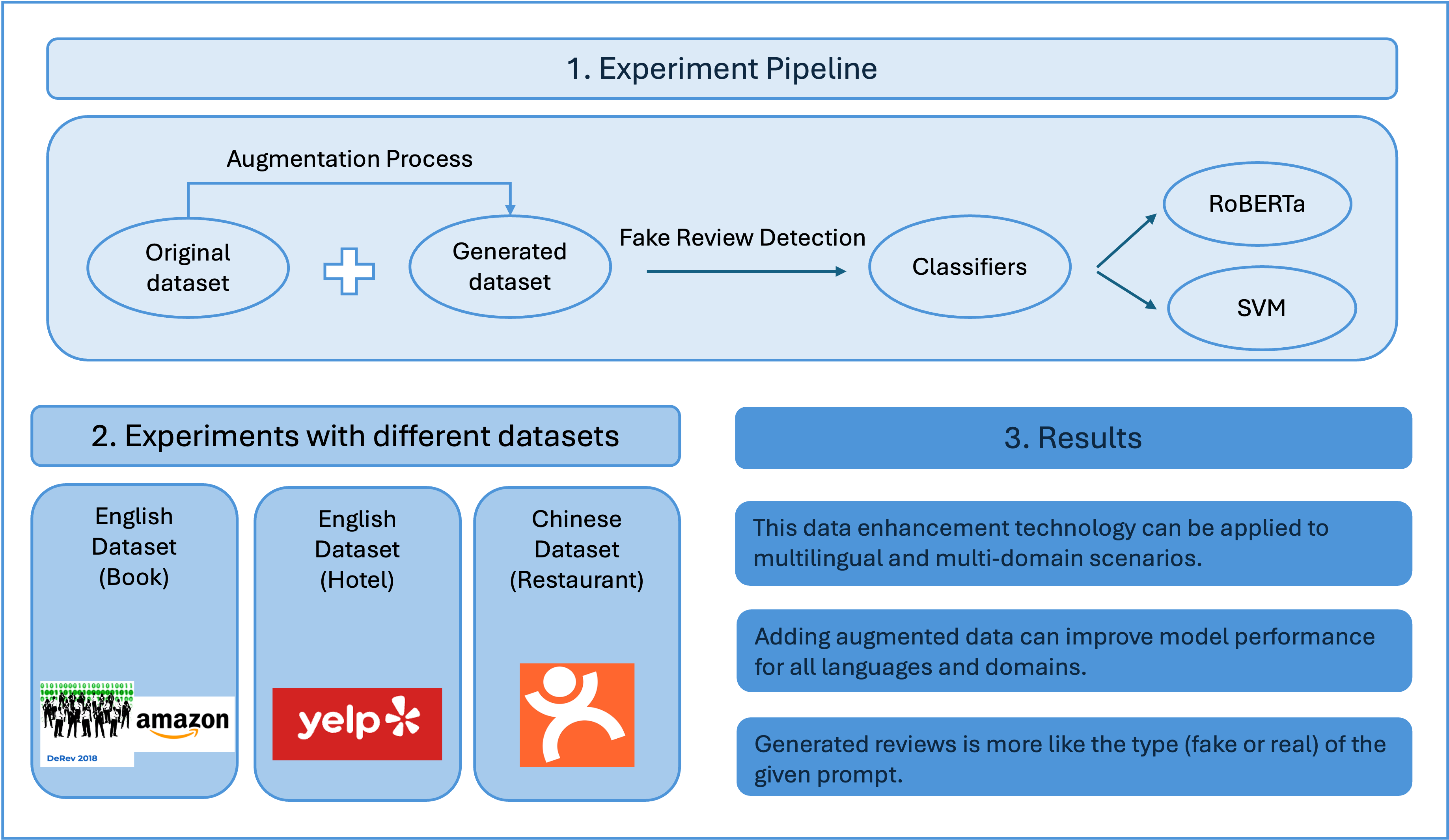}
\end{graphicalabstract}

\begin{highlights}
\item The data augmentation technique can provide similar results whether the review content is in Chinese or English and whether the review content is for books or restaurants. We observed that performance is improved in all experiment cases.

\item The authenticity of the generated dataset is related to the authenticity of the provided prompt. When fake reviews are used to generate reviews, the generated reviews tend to be more fake.
\end{highlights}


\end{frontmatter}


% \section{}
% \label{}

\section{Introduction}
\label{sec:introduction}

With the development of the Internet, users have gradually shifted from offline purchases to online purchases. 
Without being able to touch physical items, product reviews have become an important factor in deciding whether to purchase the product \shortcite{mukherjee2013fake, mohawesh2024fake, pan2024detecting}. 
These reviews provide purchasing options for subsequent customers and bring improvement directions to merchants \shortcite{mohawesh2021fake, xylogiannopoulos2024chatgpt}. 
But fake positive reviews can mistakenly lead users to purchase products that do not meet their needs. 
Fake negative reviews may lead users to abandon the product, resulting in losses for the merchant\shortcite{zhu2010impact, ho2013effects}. 
Therefore, fake product or service reviews could have a serious negative effect. 
As a result, finding and removing such fake reviews has become a topic of intensive research 
\shortcite{jindal2008opinion,ott2011finding,fornaciari2014identifying,fornaciari2020fake,fornaciari-et-al:EACL21,costa-et-al:JRCS19,vidanagama-et-al:AIR19,SALMINEN2022102771, pan2024detecting}.

Virtually all research on distinguishing real from fake reviews involves supervised methods. 
Early research used traditional machine learning methods such as Naïve Bayes, Random Forest, and Support Vector Machine (SVM)\shortcite{jindal2008opinion,mihalcea2009lie,ott2011finding,fornaciari2014identifying,crawford2015survey,maurya2024deceptive}. 
More recently, deep learning methods have become predominant \shortcite{li2015learning,fornaciari-et-al:EACL21,salminen2022creating}.
More and more scholars are using deep learning methods that do not involve hand-specified features instead of traditional methods based on review features, user behaviour or product features \shortcite{li2016detecting, wu2018hpsd, wu2020fake}.
But these methods require large datasets, which are even more difficult to acquire than for other NLP tasks, as determining which reviews are genuine and which are fake generally involves complex procedures \shortcite{mukherjee2013fake,fornaciari2020fake}.
This issue was addressed to some extent 
thanks to %through the involvement of 
large companies such as Amazon \shortcite{kaushik2018exploring}, Yelp \shortcite{rayana2015collective} and TripAdvisor \shortcite{alam2016joint}, which released versions of their data. 
However, these datasets only cover a limited range of possible types of reviews and a limited number of languages. 
For other products/services and languages, these datasets will not be of much help. 
Research such as \shortcite{fornaciari2020fake} has shown that the problem of obtaining large amounts of useful data efficiently and quickly remains a problem because crowdsourced data are very different from real data. ‘A large number of products get very few reviews and a small number of products get a large number of reviews’ This is a kind of ‘wisdom of a few’ \shortcite{fornaciari2020fake, baeza2015wisdom}.

Thus, in recent work, data augmentation methods have been proposed as a solution to the problem of creating sufficiently large datasets for fake review detection for novel products/services and novel languages which are closer to real fake reviews than 'fake reviews' produced via crowdsourcing \shortcite{aghakhani2018detecting,salminen2022creating}.

\section{Research Questions and Objectives}
\label{sec:objectives}

In this study, we propose a novel data augmentation technique for fake review detection that can be applied to multiple languages and domains.

Specifically, we address the following three questions:

\begin{enumerate}%[A]
\item Can the performance of a fake review detector be improved by adding augmented data?
\item How should the generated data be used?
\item Can the proposed data augmentation technology be applied across languages and types of reviews?
\end{enumerate}

Firstly, the objective of this research is to test whether data augmentation techniques can be a useful way to generate datasets for studying fake reviews. In cases where we only have a small amount of data, it is possible to augment the dataset with data augmentation techniques.

Secondly, we hope that the dataset obtained by data augmentation technology can be as similar as possible to the original dataset.

Thirdly, We hope that the data augmentation techniques can be wiled used across languages and domains.

\section{Background}
\label{sec:background}

\subsection{Fake Reviews Detection}

Fake reviews are reviews of products (such as books) or services (such as restaurants or hotels) that do not reflect 
the reviewer's real subjective opinion.
Thus, fake review detection is 
akin to opinion spamming detection, 
but very different from fake news detection, where objective truth is available.

Fake reviews are unfortunately widespread and are generated by users or machines without any experience with the product or service \shortcite{jindal2008opinion,lee2016capturing,pan2024detecting, hyder2024bert}.
Because fake reviews can be created manually by human writers or automatically generated by machines, technological progress will promote the advancement of fake review generation \shortcite{salminen2022creating, duma2024fake}.  \\

\noindent\textbf{Cues for Detecting Fake Reviews}

Until not so long ago, most fake review detection was done manually \shortcite{jindal2008opinion}, relying on a combination of linguistic features and meta-features known to be associated with lying \shortcite{vrji:08}.
For example, liars often use second or third-person pronouns: \textit{you}, \textit{he}, and \textit{she} to avoid using first-person pronouns to express their opinions to reduce their association with unfamiliar content \shortcite{newman2003lying,vrji:08,jindal2008opinion,mihalcea2009lie,fitzpatrick-et-al:book}. 
Meta information includes whether a user or product has very few reviews, and their reviews rarely have feedback \shortciteA{jindal2008opinion}.
\shortciteA{WALTHER2023100278} reviewed the literature on fake review detection and classified clues into five major categories. 
Review characteristics are judged by the description of the reviewed item \shortcite{ansari2021customer,peng2016consumer,filieri2016makes,jensen2013credibility}.
Textual characteristics are determined by judging the user's writing skills, such as the use of grammar \shortcite{filieri2016makes, ansari2021customer}.
Reviewer characteristics determine the authenticity of a review by judging the characteristics of the reviewing user \shortcite{deandrea2018people, kusumasondjaja2012credibility}.
Seller characteristics determine the authenticity of reviews by judging the merchant's reputation and sales record \shortcite{peng2016consumer, deandrea2018people}.
At the same time, Platform characteristics are also an important judgment indicator, such as different sales platforms and product displays \shortcite{filieri2016makes, munzel2015malicious}.
These make it easier for users to spot fake reviews online.

But  
with the explosive growth of online reviews, manual identification of fake reviews has become unfeasible. 

For example, TripAdvisor had over 200 million reviews by the end of 2015 \shortcite{crawford2015survey}. On the other hand, the accuracy of manual judgment is not very high. For example, the \shortciteA{ott2011finding} team hired humans to determine whether reviews were true or false. Its highest accuracy rate (65\%) is far lower than the accuracy rate of a machine learning algorithm (85\%). \\

\noindent\textbf{Automatic Detection Methods}

The alternative to manual deceptive review detection is using automated methods \shortcite{newman2003lying,hancock2004deception,hancock2007lying,jindal2008opinion,mihalcea2009lie,fornaciari2014identifying,fornaciari2020fake, pan2024detecting, hyder2024bert, maurya2024deceptive}.
Automatic deceptive review classification typically uses NLP tools classifying a review on the basis of keywords, N-gram and TF-IDF, etc \shortcite{newman2003lying,jindal2008opinion, mihalcea2009lie,fornaciari2014identifying,fornaciari2020fake}. 
In addition, you can also add some content that is not the main text, such as user gender, user address and the number of user reviews, etc., to improve accuracy \shortcite{crawford2015survey}. \\

\noindent\textbf{Deep Learning Methods}

With technological innovation, researchers have shifted from using traditional machine learning algorithms to using deep learning methods \shortcite{hyder2024bert}.
For example, \shortciteA{aghakhani2018detecting} proposed the FakeGAN model. The GAN network is applied to the generation and detection of fake news. It is different from the general single generator and discriminator GAN model. FakeGAN contains a generator and two discriminators. The two discriminators help the generator to better generate new data similar to the original data and also improve the accuracy of the classifier. 
\shortciteA{salminen2022creating} fine-tuned the GPT-2 model to generate a new dataset. The dataset contains ten categories of Amazon product reviews, each of which contains 2000 reviews. Then fine-tuned the RoBERTa model as a discriminator for fake review detection. \shortciteA{fornaciari-etal-2021-bertective} also tried multiple BERT-based models when studying the impact of context on fake comment detection. In addition, researchers also tend to develop new models based on deep learning algorithms. The new model improves the ability to identify online fake reviews by combining CNN with natural language processing technology and a new optimization algorithm (adaptive particle swarm optimization) \shortcite{deshai2023unmasking}. DHMFRD-TER combines BERT, CNN and LSTM models. The model's discriminative ability is further improved by adding emotion and review ratings \shortciteA{duma2024dhmfrd}.
Although the final judgment result is determined by the classification model, it always includes human effort
for 
%Including 
dataset annotation, dataset creation and data preprocessing are all done manually \shortcite{salminen2022creating}. 

\subsection{Datasets for Fake Review Detection}
% \todo{THis Section could be greatly expanded}

Datasets are needed to train fake review detection models.
%The core of fake review detection is the fake review dataset. 
Among the existing examples, we will mention datasets for 
hotel reviews \shortcite{yoo2009comparison,ott2011finding}, 
Amazon reviews \shortcite{kaushik2018exploring} and 
the DeRev2018 dataset of Amazon book reviews \shortcite{fornaciari2014identifying}. 
These datasets can be divided into three major categories: Lab-collected datasets, Crowdsourcing and Genuinely true and false review datasets. 

The first datasets for deception detection were collected by academics in the lab. 
As an example, \shortciteA{newman2003lying} distributed a series of questions in schools based on different topics and asked students to give true and false feedback. Similarly, the \shortciteA{hancock2004deception} team provided The Social Interaction and Deception form to students at a northeastern American university to obtain deception-related data.

However, this method makes it difficult to obtain the large amount of data required to train AI models. 
\shortciteA{ott2011finding} and \shortciteA{mihalcea2009lie} used a crowdsourcing platform to obtain the data.  The \shortciteA{ott2011finding} dataset contains 800 fake and 800 truthful reviews of hotel reviews. The \shortciteA{mihalcea2009lie} dataset contains 300 fake and 300 truthful reviews in three topics.

This approach does allow for the collection of large enough data. After multiple reviews and revisions, \shortciteA{fornaciari2014identifying} finally provided a dataset of 6819 highly credible fake reviews of books. This dataset includes 22 books called Innocent Books. However, the deviation between crowdsourced data and real data is difficult to ignore \shortcite{fornaciari2020fake}.

\subsection{Data Augmentation}

\noindent\textbf{Data Augmentation}

The term data augmentation refers to methods for constructing iterative optimization or sampling algorithms via the introduction of unobserved data or latent variables \shortcite{van2001art}.

In recent years, data augmentation 
has started to be widely used in the field of NLP. 
Data augmentation involves increasing the size of the training data by adding to it new synthetic data generated to resemble real data according to some criteria, while at the same time introducing diversity. 
The hypothesis is that the increased size of the training data will result in improved performance for the classifier.
For example, the sentiment analysis framework proposed by \shortciteA{shang2021campus} 
includes a transformer-based GAN model, which can generate high-quality data, thereby expanding the original training set. 
\shortciteA{junior2022use} 
used data augmentation technology when analyzing COVID-19 fake news. 
Recently, with the popularity of LLMs such as ChatGPT, researchers have also started to combine ChatGPT with data augmentation. 
One example is the ChatAug framework, 
generating auxiliary data for the few-shot text classification task \shortcite{dai2023chataug}. \\

\noindent\textbf{Data Augmentation for Fraud and Fake News Detection}

Data augmentation 
is also widely used in the field of fraud detection. 
For instance,
GAN-based data augmentation technology has 
been applied to credit card fraud detection \shortcite{make5010019}. To balance the original dataset, data augmentation techniques were used to synthesize the dataset \shortcite{make5010019}. 
\shortciteA{app13137389} 
enhanced a fake news dataset in the Romanian language using back translation and a method they called Easy Data Augmentation,
achieving 
improvements in AUC scores across all models tested. 
Similarly,  \shortciteA{HUA2023110125} 
first used Back Translation technology to enhance the dataset and then did fake news detection by combining text data and image data. \\

\noindent\textbf{Data Augmentation for Fake Review Detection}

In the field of fake review detection, a series of new models based on GAN networks have emerged. For example, 
FakeGAN \shortcite{aghakhani2018detecting} uses two discriminator models and one generator model. 
GANgster \shortcite{shehnepoor2020gangster} adds review rating score when generating reviews so that it generates score-correlated reviews. 
\shortciteA {salminen2022creating} expand the dataset using GPT-2. The RoBERTa model is then fine-tuned to detect fake and real. \\

\noindent\textbf{Data Augmentation Methods}

More than 100 data augmentation techniques 
were discussed 
and classified into 12 categories by \shortciteA{bayer2022survey}. 
Based on the differences in technology and NLP implementation, data augmentation technology can be classified as rule-based, interpolation-based and model-based \shortcite{feng2021survey}. 
At the same time, due to the different types of enhanced data, 
data augmentation methods 
can also be classified into paraphrasing-based, noising-based, and sampling-based \shortcite{li2022data}. Specifically, paraphrasing-based generates similar, paraphrased sentences by rewriting the original sentences while keeping the meaning of the sentences consistent \shortcite{li2022data}. Noising-based is to achieve data expansion by adding noise to the original data to reduce modifications to the original data as much as possible \shortcite{li2022data}. Sampling-based maintains the original data distribution and then generates a new data set based on the data distribution. The process is more complicated, but it can generate data that is different but similar to the original data as much as possible \shortcite{li2022data}. When using data augmentation technology, the choice is based on the actual application. 

This article uses the sampling-based method for data expansion. The purpose is to maintain the original data distribution as much as possible when there is a small amount of data and create a large number of datasets with similar content.

\section{Our Approach To Data Augmentation}

We propose to address the problem of data sparsity for fake review detection via data augmentation. In this section, our data augmentation methodology is introduced in detail and its specialization for multilingual and multi-domain use.

\subsection{Methodology}

\begin{figure*}[ht]
\center
\includegraphics[width=0.6\textwidth]{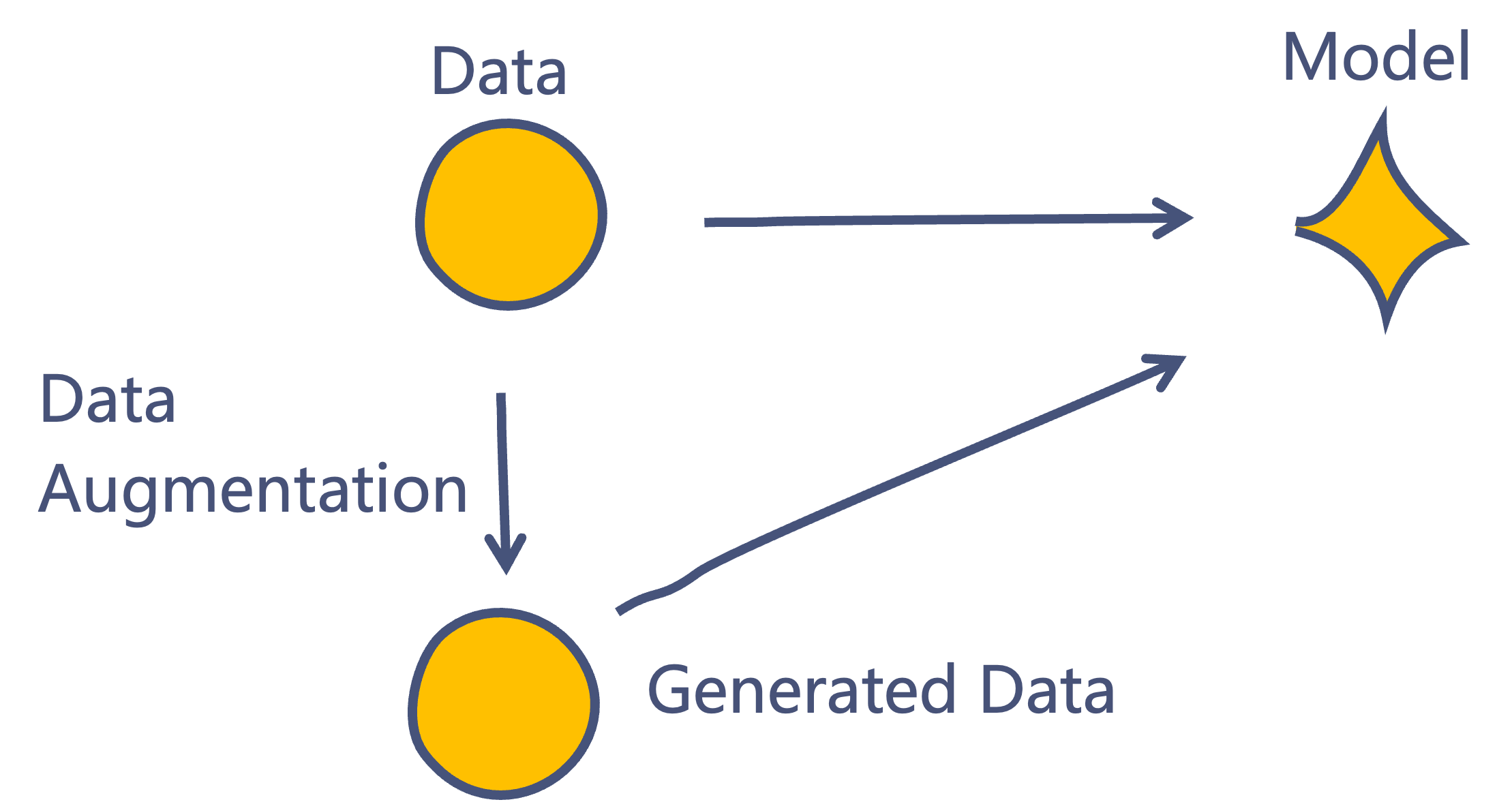}
\caption{Data Augmentation.
The new data obtained by augmenting the existing data is also used as training data for model training, thereby improving model efficiency.}
\label{fig:general-approach}
\end{figure*}

Overall, our proposed process follows the following 4 steps: 

\begin{enumerate}
    \item Generate data through a generator. Since the generated data contains different languages, the generation models will differ. 
    \item Combine the resulting data with the original data to generate different data groups. 
    \item Train two different types of fake news classifiers using SVMs and Roberta. (First, the classifier here is used as a detection tool to verify whether the model efficiency has been improved. Second, to ensure the consistency of the experiment, a newer and better classification method has not been replaced here. Using the classifier consistent with the previous experiment can more intuitively express the improvement of model efficiency.)
    \item Test the accuracy of the classifier on different data groups. 
\end{enumerate}
This approach is illustrated in Figure  \ref{fig:general-approach}. 

Because this study aims to verify whether data augmentation technology can be widely applied to multilingual and multi-domain, the data in the experiment includes comparisons between different fields in the same language and different fields in different languages.

\subsection{Generating Data: English}
To obtain the data, we created a generator through which we obtain coherent and consistent text. The simplest generator can directly use the language model to generate relevant text based on given prompts. 
In preliminary experiments, we tried using GPT-2 \shortcite{radford2019language} 
directly to generate sentences, but the results were not good.
So we started using the interpolation model proposed by \shortciteA{wang_narrative_2020} instead, achieving better results; we use this approach in this paper.

Figure \ref{fig:glm-generator-structure} illustrates the structure of our generator. For English, based on the interpolation model.
It contains a fine-tuned OPT \shortcite{opt2022zhang} text generation model and a Coherence Ranker \shortcite{moon_unified_2019} model for text selection. 
Candidate sentences are generated through the fine-tuned OPT model, and then the Coherence Ranker selects the sentence that best fits the context.

\subsection{Generalizing to Other Languages and Other Domains}

\begin{figure*}[ht]
\center
\includegraphics[width=0.8\textwidth]{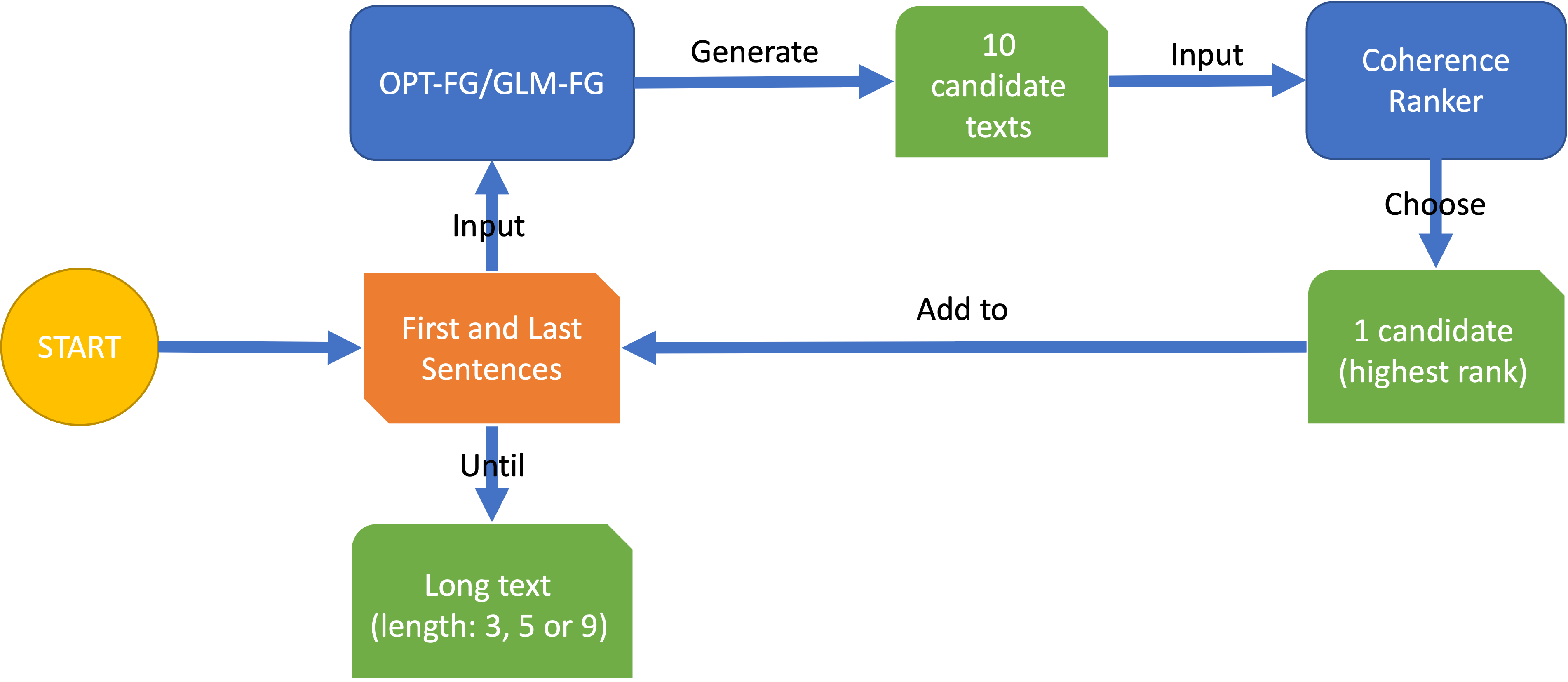}
\caption{The text generator pipeline. It contains a 
LLM %GLM-Chinese 
generator 
used to produce texts via interpolation, 
and a coherence ranker to score the produced texts. 
%and interpolation. 
It generates text of length 3, 5 or 9 after multiple iterations.}
\label{fig:glm-generator-structure}
\end{figure*}

The architecture of the model used to generate texts in Chinese is the same as that used for English and illustrated in Figure \ref{fig:glm-generator-structure}.
When conducting the experiments on DianPing TEST, because the OPT model did not support Chinese, we replaced OPT with GLM-10b-Chinese.

In addition, 
for the Chinese experiments, 
%we produced 
the GLM-Chinese-FG model was obtained by fine-tuning the GLM-10b-Chinese model on the DianPing dataset. 
GLM-10b-Chinese is pre-trained on the WuDaoCorpora dataset\shortcite{yuan2021wudaocorpora}. It has 48 transformer layers, with a hidden size of 4096 and 64 attention heads in each layer. The model is pre-trained with autoregressive blank-filling objectives designed for natural language understanding, seq2seq, and language modelling.

The 
architecture in Figure \ref{fig:glm-generator-structure} uses interpolation to obtain long, complete reviews.
The inputs are the first and last sentences of existing comments. Output 10 candidate sentences or one final sentence through the fine-tuned OPT-FG or GLM-FG model respectively. (A coherence sorter is behind a fine-tuned OPT model, then used to select the sentence that is most coherent with the input.) Loop through the entire generation process until the desired length is generated. In this experiment, we chose 5 as the review length. So the loop will only be executed three times.

\subsection{The Output of the Generator: An 
%Qualitative 
Analysis}

In this section, we 
illustrate the quality of the generation models we used via 
examples 
of the output of 
these models, 
including GPT-2, OPT and GLM (Chinese). 
We used both a quantitative metric (BLEU) and a qualitative analysis to assess the output of the models.
These generation results 
motivated 
our selection of the final model.

\subsubsection{GPT-2-Generated Text}
\noindent \textbf{Original Review (Amazon;Fake)} : 
\begin{shaded}
Upon receiving the product, the instructions were read and followed for proper 
installation. When I placed the product on a bathroom tile wall it soon fell 
off on the floor. Several attempts were made to secure it to the wall but it 
would not hold the suction and would fall to the floor. The product was 
returned for a refund at our local store.
\end{shaded}

\noindent \textbf{Review Generated by GPT-2} : 
\noindent 

\begin{shaded}
I got this for my son and he loves it! He can't believe how small, durable 
these are. I bought them as a gift but they work well enough to hold his
toddler's hand in place with no worry of slipping off when putting on the 
baby clothes or wearing any kindus bag (it is not very strong).
\end{shaded}

The review shown above was generated by GPT-2. 
Quantitatively, the automatically generated review obtains a BLEU score of 3.2947e-155. 
Qualitatively, 
this review illustrates the fact that 
when receiving the title and keywords of the review in input, GPT-2 can generally generate good review sentences, but it is often off-topic, as in  the example above, 
which should be about the shower handle. 
This led us not to use GPT-2 in our experiments - 
using instead OPT for English and GLM-10b-Chinese with context to control topics, as discussed above.

\subsubsection{OPT Generator}
\noindent \textbf{Original Review (Amazon;Real)}: 
\noindent 
\begin{shaded}
This book is a series of short stories detailing the lives of various workers 
in Iraq and Afghanistan. They live with boredom and violence in the places 
they are assigned to and then are expected to come to the US and live a 
\&\#34;normal\&\#34; life with people who have no idea of their experiences.
Very insightful.
\end{shaded}

\noindent \textbf{Review Generated by OPT (Amazon;Real)} : 
\noindent 

\begin{shaded}
This book is a series of short stories detailing the lives of various workers 
in Iraq and Afghanistan. It is a compilation of stories from various
perspectives and is more a collection of stories about the experiences of
Iraqi and Afghan workers. The author does a good job of illustrating the
challenges of working in dangerous  conditions without a lot of the details
. Very insightful. Very insightful.
\end{shaded}

The two examples above are an original review from the Amazon dataset and the corresponding OPT-generated review. 
The label Real means that the generated review was obtained by using input from a real review. 
Inputs are the first and last sentence of the original review, which is then generated by interpolation. 
The original review and the generated review have different content, but the generated review is thematically consistent given the context as input. 

Quantitatively, the OPT-generated review obtains a BLEU score of 0.2779.
Qualitatively, this example illustrates another issue: 
since the OPT model has not  'read' the book,  there is a high probability that the content of the generated review will not match the real situation, and the generated review is considered fake.  

% \todo{For English I would use cohesion / lexical diversity metrics}

\subsubsection{GLM Generator}
\noindent \textbf{Original Review (DianPing)} : 
\begin{CJK*}{UTF8}{gbsn}
\begin{shaded}
\begin{verbatim}
皮薄汤多，关键还便宜。 鲜清的很喜欢，不错，红清的烤鸭顾烤鸭山，不 鲜清的很喜欢，不错，红清的烤鸭顾烤鸭山，不烤鸭，蛋求烤鸭的鸭山，不烤鸭，鸭 服务员 据说还上过星尚的人气美食节目
\end{verbatim}
\end{shaded}

\noindent Translation: \textit{The skin is thin, the soup is rich, and the key is it’s cheap. Fresh, clear. I like the roast duck of Hongqing very much, not bad. It is said that it has even appeared on Xing Shang’s popular food program.} \\

\noindent \textbf{Review generated by GLM-10b-Chinese-RG (DianPing)} : 
\begin{shaded}
\begin{verbatim}
皮薄汤多，关键还便宜。 只要3块钱就能买一两。 每天早上排队的人都比较多。 所以感觉应该也比较符合大众口味。 据说还上过星尚的人气美食节目
\end{verbatim}
\end{shaded}
\end{CJK*}

\noindent Translation: \textit{The skin is thin, the soup is rich, and the key is it’s cheap. You can buy one for only 3 yuan. There are more people queuing up every morning. So it feels like it should be more in line with public taste. It is said that he even appeared on Xing Shang’s popular food show.}

In the original review, it said that the roast duck is cheap, fragrant, delicious, and has been on TV. By contrast, in the generated review, the specific price is described based on the first and last sentences. And based on the contextual description, "food is liked by everyone" was generated, which made it appear on TV. No specific food appears in the generated content. Therefore, the generated results are acceptable and the sentences maintain some coherence, but the content is inconsistent with the original review. BLEU score: 0.2534

\subsubsection{Summary of the Quantitative Results}

Table \ref{tab:review-metrics} shows the BLEU scores and Coherence scores of generated reviews.
The results in the Table provide further evidence that the texts produced using OPT and GLM are of better quality than the texts obtained using GPT-2.

\begin{table*}[h!]
\centering
\scalebox{0.7}{
\begin{tabular}{|c|c|c|}
\hline
\rowcolor{mygray}
Reviews & BLEU scores & Coherence Scores  \\ \hline
 GPT-2 Generated & 3.2947e-155& -14.0758 \\ \hline
 OPT Generated & 0.2779 & 0.5522 \\ \hline
GLM Generated &0.2534& 0.5313 \\ \hline
\end{tabular}
}
\caption{The BLEU score and Coherence Score of provided reviews.}
\label{tab:review-metrics}
\end{table*}

\section{Fake Review Detection Experiments and Results}
\label{section:experiments}

To answer the questions raised in section \ref{sec:introduction}, we conducted %the following 
three sets of experiments: 

\begin{enumerate}
    \item AmazonAndDeRev TEST: First of all, we used the Amazon \shortcite{SALMINEN2022102771} and DeRev \shortcite{fornaciari2014identifying} English datasets of reviews about books widely used in previous research. 
    \item Yelp TEST: Then, to further prove the usability of this data augmentation technology, we tested our approach on an english dataset for a different domain, the Yelp English dataset, which hotel reviews in North America.
    \item 
    DianPing TEST: Finally, we used the DianPing Chinese dataset to verify further that the proposed data augmentation technology can be applied to multilingual environments (and yet another domain).
\end{enumerate}

\subsection{Datasets Used}

Table \ref{tab:all-used-dataset} lists the number of real and deceptive reviews in the datasets we used.

\textbf{DEREV}
\shortcite{fornaciari2014identifying,fornaciari2020fake} consists of Amazon book reviews produced by individuals who confessed to writing fake reviews for financial gain, as well as reviews for which there is strong evidence that is genuine. 

\textbf{Amazon}
The Amazon reviews dataset contains user review data that were identified by the Amazon customer team as being clearly true or false. 
It contains 21,000 items, categorized into 30 classes, each of which contains 700 reviews.

\textbf{Yelp}
The Yelp dataset includes hotel and restaurant reviews filtered (spam) and recommended (legitimate) by Yelp.

\textbf{DianPing}
The DianPing dataset consists of all reviews of all restaurants in Shanghai from November 1st, 2011 to April
18th, 2014.

In the experiment 'AmazonAndDeRev TEST', two fake review datasets were used: the Amazon dataset used by \shortcite{salminen2022creating}and the DeRev dataset from \shortcite{fornaciari2014identifying,fornaciari2020fake}.
Both of these datasets consist of authentic fake reviews and authentic reviews;  The Amazon dataset is large and noisy, whereas the DeRev dataset is small but more high-quality.

In the experiment 'Yelp TEST', we used the Yelp dataset \shortcite{rayana2015collective}, 
which contains hotel reviews located in North America. The data includes category of user(User\_id), id of product(Product\_id), rating star(Rating), date(Date), review text(Review) and review label(Label)

In the experiment 'DianPing TEST', we used the DianPing dataset \shortcite{li2014spotting} as the experimental dataset.
This dataset contains filtered reviews identified by a fake review detection system. The data includes category label (label), user code (user), anonymous IP address (IP), rating (star) and review text (text).

In each experiment, 100\% of DianPing dataset and generated dataset are used to fine-tune the Generator models. 80\% of DianPing dataset and generated dataset are used to train the Classifier model, and 20\% of the DianPing dataset is used as a test set.

\begin{table*}[h!]
\centering
\scalebox{0.7}{
\begin{tabular}{|c|c|c|c|}
\hline
\rowcolor{mygray}
Dataset Name &Number Of Reviews & Number of  Real Reviews & Number of Deceptive Reviews  \\ \hline
Amazon Dataset & 700 & 350 & 350 \\ \hline
DeRev Dataset & 1552& 776 & 776 \\ \hline
Yelp Dataset &359053 & 322167 & 36885 \\ \hline
DianPing Dataset &9765 & 6241 & 3524 \\ \hline
\end{tabular}
}
\caption{The number of real and false reviews in used datasets.}
\label{tab:all-used-dataset}
\end{table*}

\subsection{Classifying Reviews Using the Augmented Datasets}

The purpose of the classifier experiments is to verify whether adding the generated data as discussed previously improves the performance of the model. In 
these experiments,
we used two classifiers, SVM \shortcite{boser1992training} and RoBERTa \shortcite{reborta2019}.

\subsubsection{Training Data}

The quality of the generated dataset 
can be 
judged indirectly 
through the classification performance of the classifier. 
If the classification 
performance %preference 
of the classifier with added data is better than that of the original model, which means data augmentation can improve the performance of the model. Likewise, the quality of the generated dataset is also good.

\subsubsection{The Logic of the Experiments}

\begin{figure}[ht]
\includegraphics[width=\textwidth]{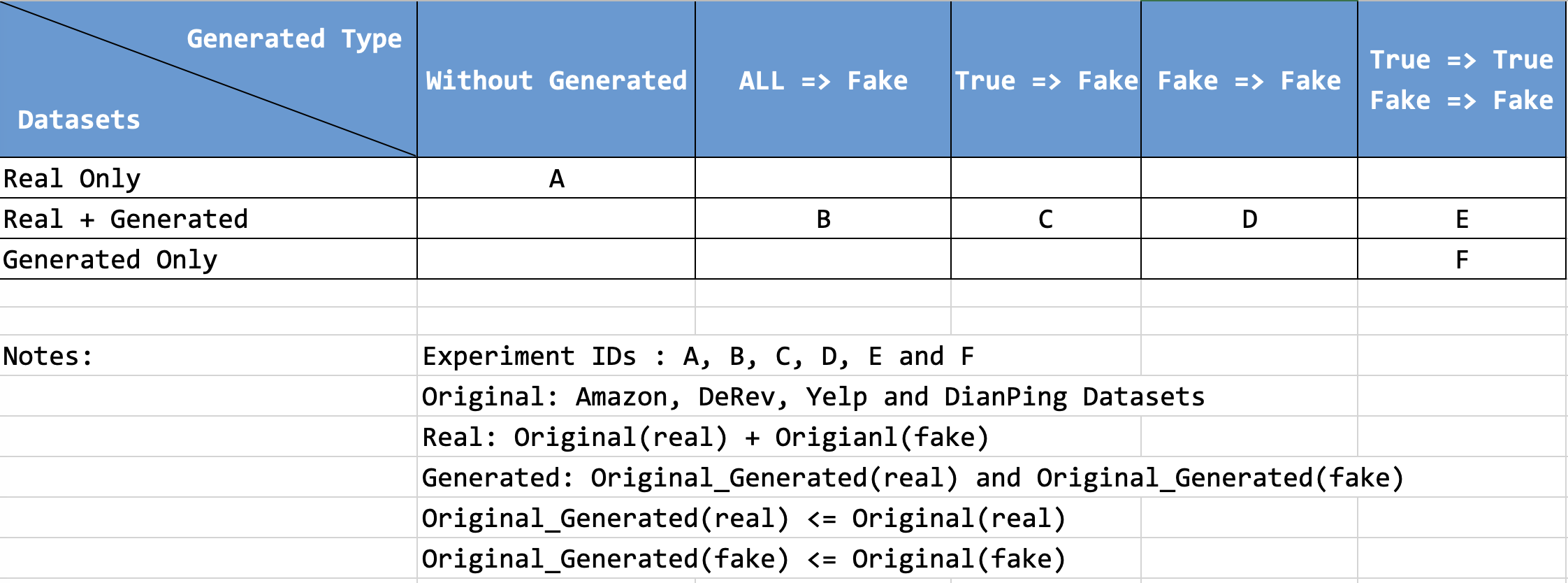}
\caption{
Summary of the training modalities. 
%The combination of datasets. 
By combining different datasets, existing models are trained and tested to verify three questions.}
\label{fig:all-experiment-logial}
\end{figure}

Our classifier experiments were designed to answer the three questions in Section \ref{sec:introduction}. 
To answer these questions, we ran the series of experiments shown in Figure \ref {fig:all-experiment-logial}. 

To answer Question 1 (\textit{Can the performance of a fake review detector be improved by adding augmented data?}), 
we carried out experiments A and E.

In Experiment A, only the original dataset 
is used for training. 
Experiment E adds the generated dataset to Experiment A.
In addition, we also ran experiment F in which only generated data were used, to judge the quality of the generated data.

Notice that the datasets contain two types of reviews:
real reviews (Original(real)) and fake reviews (Original(fake)). The generated reviews can also be divided into two types: 'real generated reviews' (Original\_Generated(real)), i.e., reviews generated from real reviews;
and 'fake generated reviews' (Original\_Generated(fake)), i.e., reviews generated from fake reviews.
This observation leads to Question 2: \textit{How should the generated data be used?}
To answer this question, 
we conducted experiments B, C and D. 
In Experiment B,  all generated reviews are considered fake.
In Experiment C only  "real generated data" --i.e., reviews generated from real reviews--are considered fake. 
In Experiment D, conversely, only  'fake generated data'--i.e., reviews generated from fake reviews--are considered fake.

\subsubsection{Book Review Classification in English (AmazonAndDeRev TEST)}

%\todo{Again, I would just discuss the results here without mentioning the previous paper}

In this first set of experiments, we used two datasets of book reviews in English, the Amazon dataset and the DeRev dataset. 
%\shortcite{liu-poesio-2023-data}. 
Empirically, we found that whereas adding DeRev training to Amazon training improved the performance on Amazon TEST, the other way around wasn't true, so we tested our models separately on Amazon TEST and DeRev TEST.

The instantiations of the training configurations A-G used for testing with DeRev test dataset are as follows:

\begin{enumerate}[A]
\item DeRev 
\item DeRev + Amazon
\item DeRev + Amazon + Generated(from Amazon(real + fake))
\item DeRev + Amazon + Generated(from Amazon(real + fake) as fake)
\item DeRev + Amazon + Generated(from Amazon(real) as fake)
\item DeRev + Amazon + Generated(from Amazon(fake) as fake)
\item DeRev + Generated(from Amazon(real + fake))
\end{enumerate}
There is also G\_Balanced which is based on Experiment G. The dataset is balanced based on Experiment G.

For Amazon TEST, we specified training configurations A-L  used for testing with Amazon test dataset as follows:

\begin{enumerate}[A]
\item DeRev
\item DeRev + Amazon
\item DeRev + Amazon + Generated(from Amazon (real + fake))
\item DeRev + Amazon + Generated(from Amazon(real + fake) as fake)
\item DeRev + Amazon + Generated(from Amazon(real) as fake)
\item DeRev + Amazon + Generated(from Amazon(fake) as fake)
\item DeRev + Generated(from Amazon (real + fake))
\item Amazon
\item Amazon + Generated(from Amazon (real + fake))
\item Amazon + Generated(from Amazon(real + fake) as fake)
\item Amazon + Generated(from Amazon(real) as fake)
\item Amazon + Generated(from Amazon(fake) as fake)
\end{enumerate}

In the experiments DeRev TEST and Amazon TEST, we used the DeRev dataset as our basic dataset. Because the DeRev dataset is of high quality and is more conducive to model training. It is more efficient to verify whether the model has been improved by adding the original or more generated Amazon dataset.

\subsubsection{Hotels and Restaurants Reviews Classification in English (YELP TEST)}

To address Question 3 (\textit{Can the proposed data augmentation technology be applied across languages and types of reviews?}) we applied our approach to two other datasets.

The Yelp dataset is still in English, but in a different domain: reviews of hotels and restaurants. 
Training configurations  A-F for Yelp TEST are shown below, consistent with the figure \ref{fig:all-experiment-logial}.
\begin{enumerate}[A]
\item Yelp
\item Yelp + Generated(from Yelp(real + fake) as fake)
\item Yelp + Generated(from Yelp(real) as fake)
\item Yelp + Generated(from Yelp(fake) as fake)
\item Yelp + Generated(from Yelp(real + fake))
\item Generated(from Yelp(real + fake))
\end{enumerate}

This dataset combination is consistent with the dataset combination of DianPing TEST described below. For specific parameter settings, please refer to DianPing TEST.

\subsubsection{Food Reviews in Chinese (DianPing TEST)}

Finally, DianPing was used to test our proposed approach to a different language (and yet another domain).

\noindent 
The full list of variants of training datasets used in the experiments  on DianPing is as follows:
\begin{enumerate}[A]
\item Dianping
\item Dianping + Generated(from DianPing(real + fake) as fake)
\item Dianping + Generated(from DianPing(real) as fake)
\item Dianping + Generated(from DianPing(fake) as fake)
\item Dianping + Generated(from DianPing(real + fake))
\item Generated(from DianPing(real + fake))
\end{enumerate}

80\% of the DianPing dataset is treated as the train set; 20\% of the DianPing dataset is treated as the test set. 

\subsubsection{Results}

\begin{figure}[ht]
\includegraphics[width=\textwidth]{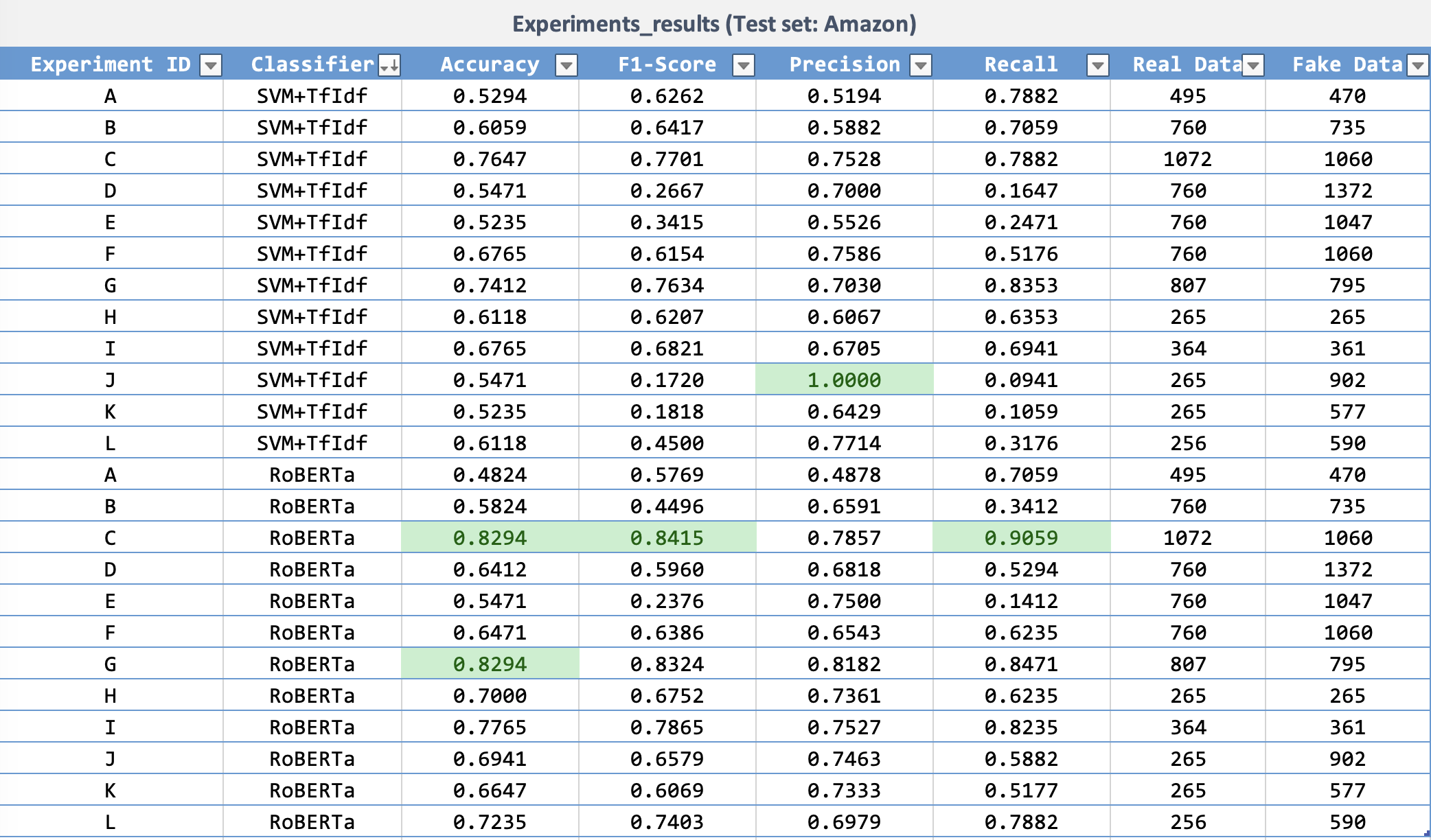}
\caption{The results for Amazon TEST}
\label{fig:amazon-test-result}
\end{figure}

\begin{figure}[ht]
\includegraphics[width=\textwidth]{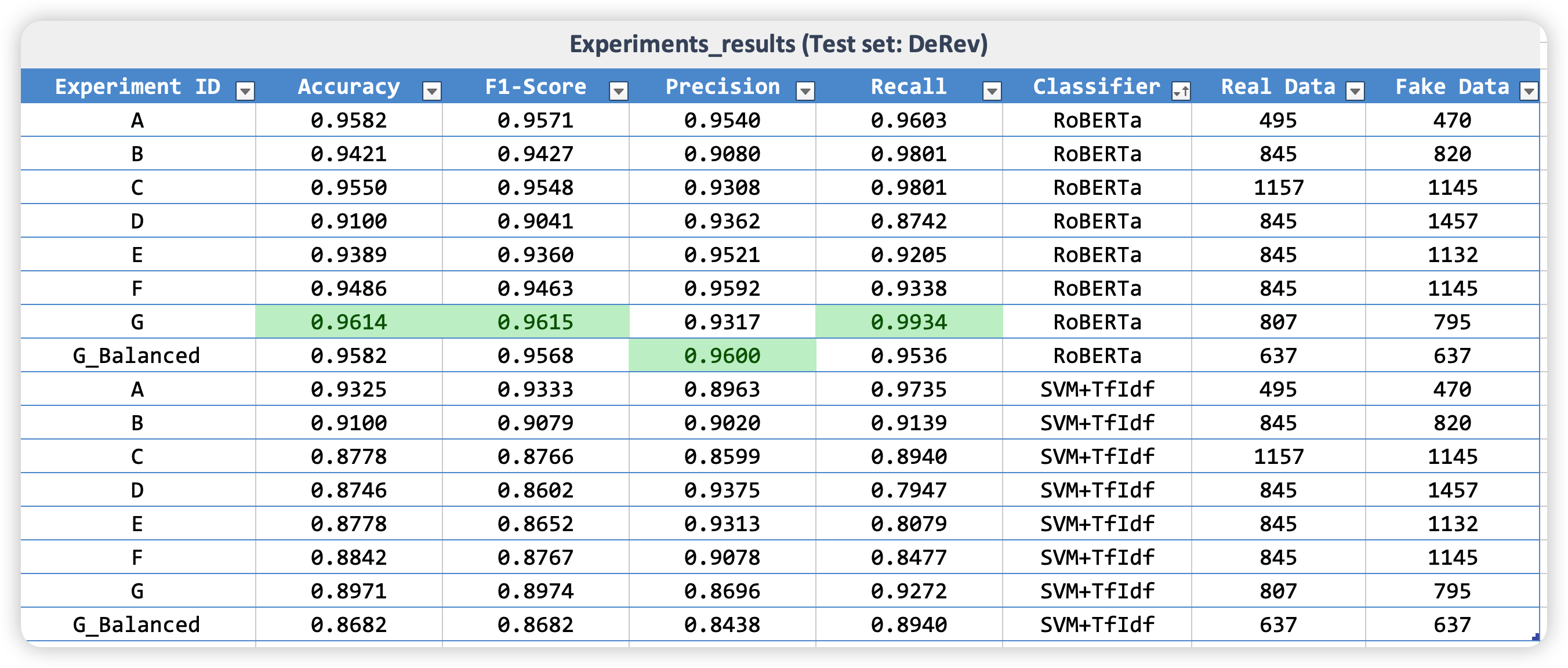}
\caption{The results for DeRev TEST}
\label{fig:derev-test-result}
\end{figure}

Figure \ref{fig:amazon-test-result} and \ref{fig:derev-test-result} show the results with Amazon TEST and DeRev TEST respectively.

\begin{figure}[ht]
\includegraphics[width=\textwidth]{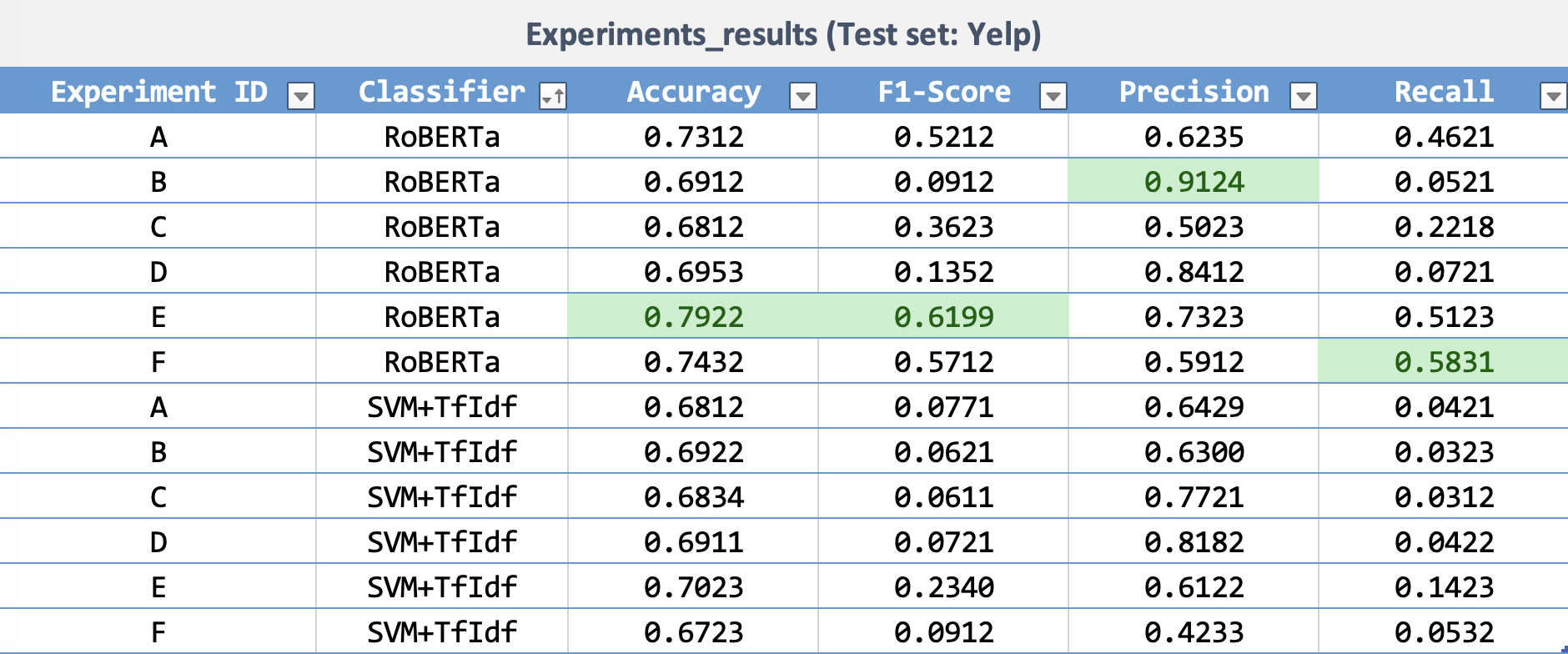}
\caption{The results for Yelp TEST}
\label{fig:yelp-test-result}
\end{figure}

\begin{figure}[ht]
\includegraphics[width=\textwidth]{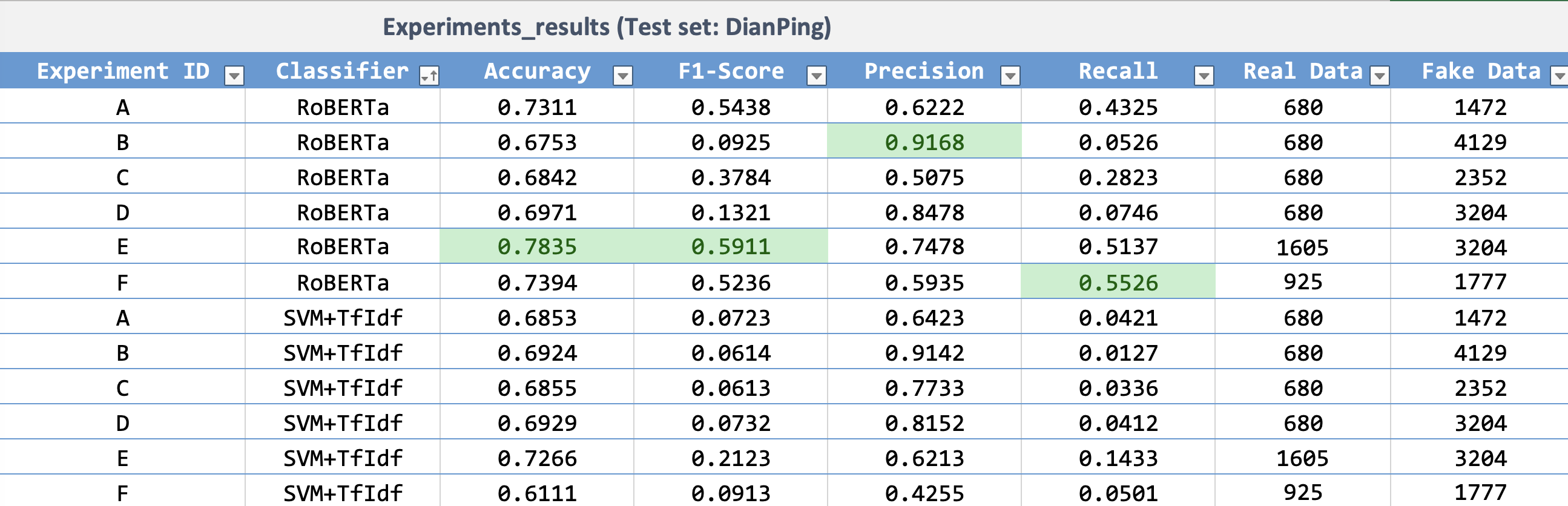}
\caption{The results for DianPing TEST}
\label{fig:dianping-test-result}
\end{figure}

These results show, first of all, that the performance of the SVM model is always lower than that with RoBERTa, throughout figure \ref{fig:yelp-test-result} and \ref{fig:dianping-test-result}. 
Therefore, the discussion below will focus on the RoBERTa model.

Secondly, 
the results show that using data augmentation,  an accuracy improvement of up to 0.31 and 7.65 percentage points can be achieved on Amazon TEST and DeRev TEST, respectively.

We found that the results of Yelp TEST and DianPing TEST are very similar, so only the results of DianPing TEST are discussed below.

Figure \ref{fig:dianping-test-result} illustrates the result with DianPing Test.  
First of all, we find that the accuracy of training configuration E is slightly higher than those obtained with training configurations A, B, C, D and F.

Comparing Experiment B with Experiments C and D shows that the generated datasets are also more similar to their corresponding categories. 'Fake dataset' generated from fake data is more like a fake dataset. Likewise, a 'Real dataset' generated from a true dataset is more like a real dataset. In addition to this, we can also distinguish between fake and real data in this way.

The results of experiments A and E show that adding additional datasets can improve the performance of the classifier. 
Finally, the experimental results show that adding a refined dataset can improve the performance of the classifier. But it didn't improve much.

\section{Discussion}

In this Section, we discuss the results of our experiments, starting by revisiting the research questions raised in section \ref{sec:introduction}.

%In this section, we compare the improvement efficiency of several groups of TEST and answer the three research questions raised in section \ref{sec:introduction}

\subsection{Answers to Research Questions}

%\todo[inline]{Why do you only discuss Yelp and DianPing here?}
\noindent \textbf{Question 1: Can model performance be improved by adding augmented data?} \\
Answer: Results based on AmazonAndDeRev TEST, Yelp TEST and DianPing TEST show that adding augmented data can improve model performance. Even without original data, only augmented data can slightly improve model performance. \\

\noindent \textbf{Question 2: What to do with the generated dataset?} \\
Answer: Results based on AmazonAndDeRev TEST, Yelp TEST and DianPing TEST show that the type of generated reviews is more like the type (fake or real) of the given prompt. \\

\begin{figure}[ht]
\includegraphics[width=\textwidth]{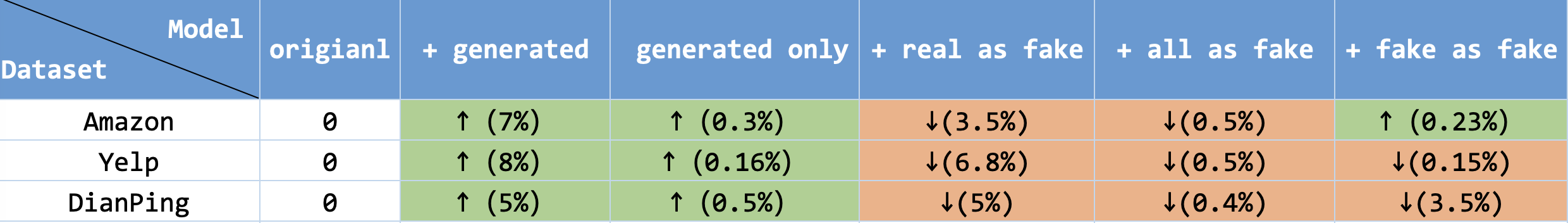}
\caption{Comparison Between Amazon Test and DianPing Test}
\label{fig:comparison-amazon-yelp-dianping}
\end{figure}

\noindent \textbf{Question 3: Can this data augmentation technology be applied to multilingual and multi-domain?} \\
Answer: Figure \ref{fig:comparison-amazon-yelp-dianping} compares Amazon TEST, Yelp TEST and DianPing TEST. Except for the last column, the improvement rates compared to the baseline are similar. When the generated data was added, the accuracy of the model increased by 7\%, 8\% and 5\% respectively. Even when using only generated data, the model accuracy is slightly improved compared to the baseline. The comparison results based on Yelp TEST, Yelp TEST and DianPing TEST show that this data enhancement technology has generalization, that is, it can be applied to multilingual and multi-domain scenarios.

\subsection{General Discussion}

Through experiments, we provide a way to generate fake review data. We also conducted some experiments on this basis. The experimental results show that we can control the authenticity of the generated data. At the same time, if we add watermarks to the generated data, we can better control the generated data to prevent other users from directly identifying the data set generated by this method as real data.

The results of the experiments also pointed out that data after data augmentation can improve the accuracy of models. It provides other researchers with ideas for improving the model. By expanding the original data set through data augmentation technology without changing the training model, the model efficiency can also be improved. Although at this stage, the length and format of the data set generated by the model are unified, it can still improve the model. In the future, it is also a feasible improvement to set the model generation results so that it generates datasets closer to that collected by humans.

\subsection{Theoretical Implications}

The main theoretical implication of our experiments is that the data generation method we introduce
has universal applicability. 
%is universal. 
Adding the generated data to training 
results in improvements in performance at fake review detection even if
the original data come from multiple languages and multiple domains. 
%The generated data can also effectively improve the performance of the model.

At the same time, we also found a correlation between the original data and the generated data. When using fake reviews to generate new reviews, the generated reviews are 
more similar to %more likely to be 
fake reviews. 
And vice versa.

\subsection{Practical Implications}

Merchants use new technologies to improve discriminators' ability to identify fake reviews. Illegal operators improve the reviews generated by improving the generator. This competition is difficult to avoid, but from the perspective of researchers, we prefer to improve the performance of the discriminator. To do this, the discriminator model needs to be updated all the time. The training model uses the latest dataset and updates the discriminator component. The idea is similar to online learning, constantly feeding the model new datasets to improve model performance \shortcite{cardoso2018towards}. When data augmentation techniques are applied, discriminator model developers will greatly save time in gathering datasets and can update models efficiently.

In general, researchers are using new technologies to improve the accuracy of discriminator models. Helping customers to more efficiently distinguish between high-quality and low-quality product reviews can increase users' confidence in merchants and even shopping platforms.

The dataset generated using this method can have a variety of effects. Merchants can use the generated positive fake reviews to influence users' desire to buy their products \shortcite{salminen2019machine}. Similarly, merchants can also use the generated negative fake reviews to influence users to buy the products of rival merchants.

Although we have created this generation pipeline, it is very simple for researchers with a certain computer-related foundation to fine-tune a suitable topic-related generation model. Therefore, there will be some illegal users who use this technology for illegal operations. This is very fatal and also a risk of computer text generation technology.

\subsection{State-Of-the-Art Comparison}

In this paper, we have shown that it is possible to improve fake review detection using data augmentation even without using the latest LLMs. We expect that the results will be improved even more using a more recent LLM, that is expected to produce text of higher quality. 

No model has been tested on all our datasets and could therefore serve as a baseline, but in the dataset for which a comparison is possible,  DeRev TEST, our model achieves F1=96.15 whereas the previous state-of-the-art \shortcite{fornaciari2020fake} was F1=92.9.

% \section{Related work}

% To improve performance on 
% %To better solve the problem of 
% fake review detection, 
% researchers are increasingly using multimodal methods--
% using not only text, but text and emotion, or text and images.
% %A single review text input does not provide more comprehensive information. So using text and emotion, and even images, is becoming more and more popular.
% One example is adding review scoring when doing fake review detection \shortcite{app13137389, 10105368}. 
% %In other efforts, 
% %Or add 
% Relevant picture information was added by \shortciteA{HUA2023110125}. 
% \todo[inline]{I don't understand the following}
% The BSTC model \shortcite{ electronics12102165} 
% uses sentiment knowledge in %Enhanced 
% pre-training to obtain emotional features, and BERT to obtain contextual sentiment features.
% The  two features above are then combined as the input of TextCNN, and finally a new feature is obtained as the input of the classifier \shortcite
% { electronics12102165}.

\section{Conclusions}

Fake review detection research is hampered by the lack of suitable resources. 
Previous research demonstrated that additional data created using crowdsourcing are substantially different from genuine fake reviews \shortcite{fornaciari2020fake}.
In this paper, we showed that the data augmentation method proposed in this paper can improve model performance for two different languages and three different domains. 
We also showed that our method results in reviews whose style resemble that of the seed data--when genuine fake reviews (produced by spammers) are used to generate synthetic reviews, the generated reviews tend to resemble these fake reviews more than genuine authentic reviews. 

% In conclusion, the data augmentation technique can provide similar results whether the review content is in Chinese or English and whether the review content is for books or restaurants. All can improve the model effect to a certain extent.

% What's more, the authenticity of the generated dataset is related to the authenticity of the provided prompt. When fake reviews are used to generate reviews, the generated reviews tend to be more fake.



% \bibliographystyle{elsarticle-harv}

\bibliographystyle{apacite}
\bibliography{mybibs}





% \end{thebibliography}
\end{document}